\documentclass[letterpaper, 10 pt, conference]{ieeeconf}  % Comment this line out
\IEEEoverridecommandlockouts                              % This command is only
\usepackage{graphicx} % for pdf, bitmapped graphics files
\usepackage{caption}
\usepackage{subcaption}
\usepackage{float}
\usepackage{amsmath} % assumes amsmath package installed
\usepackage{amssymb}  % assumes amsmath package installed
\usepackage{color}
\usepackage{multirow}
\usepackage{longtable}

\title{\LARGE \bf
Compressed Sensing for Tactile Skins
}

\author{Brayden Hollis, Stacy Patterson, and Jeff Trinkle% <-this % stops a space
\thanks{This work was partially supported by NSF Grant CCF-12084 (through the National Robotics Initiative), NSF Grant CNS-1527287, and the NSF Independent Research and Development Program. B. Hollis is supported by a SMART Scholarship, funded by OASD/R\&E (Office of the Assistant Secretary Defense-Research and Evaluation), Defense –Wide / PE0601120D8Z National Defense Education Program (NDEP) / BA-1, Basic Research. Any opinions, findings, and conclusions or recommendations expressed in this material are those of the authors and do not necessarily reflect the views of the funding agencies.}% <-this % stops a space
\thanks{B. Hollis, S. Patterson, and J. Trinkle are with the Department of Computer Science, Rensselaer Polytechnic Institute,
110 8th Street, Troy, NY, USA,\tt{ hollib@rpi.edu,\{sep,trink\}@cs.rpi.edu}.}
}

\newcommand{\measMat}{\ensuremath{\Phi}}
\newcommand{\basis}{\ensuremath{\Psi}}
\newcommand{\mR}{\mathbb{R}}
\newcommand{\ev}{\ensuremath{\epsilon}}

\begin{document}

\maketitle
\thispagestyle{empty}
\pagestyle{empty}

%%%%%%%%%%%%%%%%%%%%%%%%%%%%%%%%%%%%%%%%%%%%%%%%%%%%%%%%%%%%%%%%%%%%%%%%%%%%%%%%
\begin{abstract}
Whole body tactile perception via tactile skins offers large benefits for robots in unstructured environments. To fully realize this benefit, tactile systems must support real-time data acquisition over a massive number of tactile sensor elements. We present a novel approach for scalable tactile data acquisition using compressed sensing. We first demonstrate that the tactile data is amenable to compressed sensing techniques. We then develop a solution for fast data sampling, compression, and reconstruction that is suited for tactile system hardware and has potential for reducing the wiring complexity. Finally, we evaluate the performance of our technique on simulated tactile sensor networks. Our evaluations show that compressed sensing, with a compression ratio of 3 to 1, can achieve higher signal acquisition accuracy than full data acquisition of noisy sensor data.

%
%Whole body tactile perception offered by tactile skins offer large benefits for robots in unstructured environments. To fully realize this benefit, data acquisition must greatly improve for such systems. This paper presents an novel method for addressing this problem though compressed sensing. This paper presents the specific requirements for applying compressed sensing; representation basis, measurement matrix, and reconstruction algorithm.  It then provides suitable solutions for these components and verifies this through simulation experiments.

\end{abstract}

%%%%%%%%%%%%%%%%%%%%%%%%%%%%%%%%%%%%%%%%%%%%%%%%%%%%%%%%%%%%%%%%%%%%%%%%%%%%%%%%
\section{INTRODUCTION}

With the advances in robot technology, there is an increased desire to move robots from controlled environments to unstructured human environments. The National Robotics Initiative states that part of its goal is ``to accelerate the development and use of robots in the United States that work beside or cooperatively with people''~\cite{nri}. To achieve this goal, robots require improved perception to safely handle the uncertain and dynamic structure of these environments. Improved perception becomes even more important as robots need to effectively manipulate items in the environment. Tactile sensors, sensors that gather data through contact (\emph{e.g.}, force and temperature), are an active research area with high potential to help address this issue.

An important area of research in tactile sensing is tactile systems that cover large portions of robot bodies.  These systems, often called \emph{tactile skins}, potentially provide improvements in a robot's awareness and manipulation abilities.  For example, Ohmura et al.~\cite{box} demonstrate a robot lifting a heavy box with use of a tactile skin, a task beyond the robot's capability with traditional methods. A number of other tasks, especially for complex robots like humanoids, would benefit from tactile sensors on the surfaces of various links.

One of the challenges with developing tactile skins is the data acquisition process. It has been suggested that to achieve human-like perception, tactile systems should have spatial resolution up to 1~mm~\cite{dahiya10tactile}.  This translates to on the order of tens of thousands to millions of individual sensing elements, known as taxels, for human adult-sized robots (currently the largest system contains approximately 4200 taxels~\cite{icub}). Full utilization of the data involves integrating it into control loops.  This requires real-time acquisition and processing rates on the order of 1~kHz (\emph{e.g.}, for high-fidelity force control). In addition, due to the distributed nature of the system, a physical network is required for data transfer but is limited by physical requirements (\emph{e.g.}, weight, space, etc.)~\cite{dahiya10tactile}. Finally, sensors are typically noisy, which can further affect the acquisition process. These combined requirements and constraints pose a challenge to acquiring the data from tactile skins.

There have been multiple proposed and implemented tactile skins from the research community. Most use a data acquisition process, which we refer to as full data acquisition, of polling all the sensors directly~\cite{ohmura06sensor, schmitz11icub, mitt11hex}, relying on fast hardware for high sampling rates.  These systems typically have a subset of local sensors that are read serially via a micro-controller which then passes the data on a high capacity network it shares with other micro-controllers. The number of sensors that can be associated with a micro-controller is limited.
The number of micro-controllers connected to the network is also limited to maintain the desirable high sampling rate. To scale this approach to full body tactile skins will require multiple of these high capacity networks, which is considered undesirable~\cite{dahiya10tactile} due to physical constraints.  This method has been implemented with a sampling rate of up to 100~hz (though the number of sensors is unclear as this value is given for the local level sampling)~\cite{schmitz11icub}.

An alternate approach to full data acquisition is to use local processing of sensor data within small clusters of sensors~\cite{riman,schmitz11icub}.  This approach speeds up the entire tactile data processing rates through parallel local processing and only transferring aggregate information. Local processing like this suffers from a loss in data quality or a reduction in the versatility of the system as features must be determined \emph{a priori}. Full data acquisition, on the other hand, has the freedom to extract the various relevant features for many different applications. Local processing has been implemented with sampling rates up to 500~hz~\cite{schmitz11icub}.
 
We propose to address the tactile data acquisition problem using compressed sensing~\cite{cs}. Compressed sensing is a popular technique for signal processing in which the signal is simultaneously sampled and compressed. Provided the original signal is sparse in some basis, it can be efficiently reconstructed from this compressed version with little to no loss.  In this work, we demonstrate the feasibility of a compressed sensing-based tactile data acquisition approach using existing compressed sensing tools.
We first show that tactile sensor data is amenable to compressed sensing by identifying a basis under which the data is sparse.  We then identify sampling and compression operators that are suited for the tactile system hardware.  Finally, we explore different methods for real-time signal reconstruction from the compressed data.

Our approach shares the benefits of both the approaches previously discussed and overcomes the short comings. Similar to the local processing approach, our approach reduces the amount of data to be transferred, allowing for faster sampling rates, but it also allows the system to have the full data set and the resulting flexibility for multiple applications, like the full data acquisition approach.  Our work also has potential to reduce wiring complexity as the sampling and compression is amiable to hardware implementations that share wires across multiple taxels.

The rest of the paper is organized as follows. Section~\ref{sec:prob} describes the particular problem we address and the data set we used. We then describe compressed sensing and compress sensing requirements related to tactile skins in Section~\ref{sec:bkgd}. Our proposed solution is discussed in Section~\ref{sec:soln} and is evaluated in Section~\ref{sec:rslt}. The paper then finishes with concluding remarks and future work in Section~\ref{sec:concl}.

%%%%%%%%%%%%%%%%%%%%%%%%%%%%%%%%%%%%%%%%%%%%%%%%%%%%%%%%%%%%%%%%%%%%%%%%%%%
\section{PROBLEM SETTING}\label{sec:prob}

\subsection{Problem Description}
We address the problem of real-time data acquisition from a large-scale array of taxels. 
The array consists of $N$ individual taxels arranged in a $\sqrt{N} \times \sqrt{N}$ flat grid.  
In this paper, we consider tactile skins whose elements each output the magnitude of the contact force normal to the plane of the array at its location. However, we believe our approach can be extended to other tactile sensors that measure one-dimensional data, such as temperature.

We represent the sensor data as a vector $x(t) \in \mR^N$, where each element, $x_i(t)$, corresponds to the force applied to $i^{th}$ taxel at time-step $t$. 
The sensors do not measure the applied force exactly, but rather, they measure $x(t) + \epsilon(t)$, where $\epsilon(t) \in \mR^N$ is the sensor noise.
The tactile data acquisition problem is to efficiently transfer $x(t)$ (or a close approximation of it) from the sensors to the central processing unit.  
%These transfers are the definition of a time-step.  
We note that ``efficiently'' is a subjective term that depends on the system and application. Ideally, we aim for solutions that require on the order of milliseconds
so that the sensed data can be used in real-time feedback control.  Additionally, we desire solutions that reduce the network hardware complexities.
%"Efficiently" incorporates speed, ideally a transfer time on the order of milliseconds or less, and resource utilization. In hardware resource utilization relates to space for wires, costs, power, etc., but for this paper we generalize to amount of network wiring/number of interactions between sensors (less is preferable) as this generalizes to multiple systems and is a major component for resource utilization.

As discussed in the introduction, approaches that directly collect sensed data suffer from scalability limitations.
We consider an alternate approach based on the new and popular technique of compressed sensing~\cite{cs}.
 In compressed sensing, the data is compressed in hardware during the acquisition process.  This compressed data can be more efficiently transmitted to the central processing unit, since the reduced data requires fewer high capacity networks. At the central processing unit,
 the original signal can be efficiently recovered.  This unique approach will enable scaling tactile skins to large numbers of sensors due to the data reduction and the hardware compression techniques that allow simpler wiring.

\subsection{Data Set Description}
For our study, we used the Gazebo 3.0~\cite{gazebo} simulator with the SkinSim 0.1.0 simulation environment~\cite{habib2014skinsim} to generate data sets.
SkinSim 0.1.0 models a tactile skin as a layer of spheres individually attached to a surface by a spring and damper.  SkinSim is currently limited to planar grid arrays. 
We set up our system with each sphere-spring-damper system representing a single taxel with a sphere radius of 2.5~mm. We had no space between taxels.
We generated sensor data using two different sensor array sizes, $40 \times 40$ and $64 \times 64$, which matches the number of taxels in the current largest system.  Each sensor measured contact normal force values between 0 and 2.5 Newtons.

We use four different scenarios to demonstrate that our approach works with contact at different scales (\emph{i.e.}, number of active sensors), shape complexities, and locations on the grid.  In the first scenario, LIFT BOX, we used the 10~cm wooden cube from Gazebo's default models.
We dropped the object onto the sensor pad, with the face parallel to the pad.  After the object settled, it was lifted from the pad with constant velocity of 0.1~$\frac{\text{m}}{\text{s}}$.
The second scenario, LIFT WRENCH, followed the same procedure, but used Gazebo's default model for a monkey wrench.
In the third scenario, PATH, a metal peg, which was forced to remain vertical throughout the scenario, was dropped in a random position on the sensor pad and then moved along a random path parallel to the plane of the tactile array.  
The final scenario, DRAG WRENCH, again used the monkey wrench.  After dropping onto the sensor pad, the wrench was moved with constant velocity of 0.05~$\frac{\text{m}}{\text{s}}$ also parallel to the sensor plane.
The sensors' heights caused them to rest above the ground plane, so, as the wrench moved off the sensor pad, it was partially supported by the lower ground plane.
All of the scenarios except PATH were run until the object stopped making contact with the sensor array.  PATH had no guarantees of leaving the sensor pad, so it was run for an extended period of time.

%Note we only used 2 scenarios at with the $64 \times 64$ array.
%
%Our data collections was done over four scenarios with the first two repeated using the larger sensor grid.  All the scenarios start by dropping the object onto the sensor pad, letting it settle and then manipulating it for the scenario.  Our first scenario used the 10cm wooden cube from Gazebo's default models \note{cite?}.  After it was dropped on the sensor pad, with the face parallel to the pad and stabilized, it was lifted from the pad with a constant velocity.  Our second scenario followed the same procedure but used the default monkey wrench instead of the cube. The third scenario used a metal peg which was forced to remain vertical throughout the scenario.  It was dropped in a random position on the sensor pad and then was moved along a random path.  The last scenario used the monkey wrench again, but after dropping, it was given a constant velocity parallel to the sensor plane.  As the sensors height cause them to rest above the ground plane, as the wrench moves off the sensor pad, it is partially supported by the lower ground plane.

%For all of our simulations, we noticed each time step the sensor values were oscillating, even with different timesteps.  As this is an unrealistic situation, we corrected by smoothing the data with a moving window average with a window size of 10.
For the experiments on the $40 \times 40$ and $64 \times 64$ arrays, the simulation time-steps were chosen to be 1 and 0.1 milliseconds respectively. Table~\ref{tab:Data} shows the number of time-steps in which we collected data for each scenario.  We set high sampling rates as dynamics simulators perform better with small time-steps.
We used all the sampling data since we are more interested in a sequence of tactile data than the specific scenario.  So, while the simulator was set up with the object falling under gravity, due to realistic tactile sampling rates, we treat the scenario more as the object being manipulated against the sensors with a slower velocity. 
Even with the small time-steps, the data generated by the SkinSim simulator exhibited high-frequency oscillations in the contact force signals.  We believe this is due to integration error of the physics engine or to low damping.  We compensated for this physically unrealistic behavior by 
applying a moving average filter with a width of 10 time-steps.

\begin{table}[ht]
\caption{Number of simulation time-steps of data for each scenario}
\begin{center}
  \begin{tabular}{| c | c | c | c | c | c |}
    \hline
    Scenario    & Array & number of time-steps \\ \hline
    LIFT BOX    & $40 \times 40$ &   700 \\ 
    LIFT WRENCH & $40 \times 40$ &  3000 \\
    DRAG WRENCH & $40 \times 40$ &  5000 \\
    PATH        & $40 \times 40$ & 12200 \\
    LIFT BOX    & $64 \times 64$ &  7600 \\
    DRAG WRENCH & $64 \times 64$ & 12600 \\
    \hline
  \end{tabular}
\label{tab:Data}
\end{center}
\end{table}

%%%%%%%%%%%%%%%%%%%%%%%%%%%%%%%%%%%%%%%%%%%%%%%%%%%%%%%%%%%%%%%%%%%%%%%%%%%%%%%%
\section{BACKGROUND on COMPRESSED SENSING}\label{sec:bkgd}

\subsection{Compressed Sensing Theory} \label{sec:cstheory}

Let $x \in \mR^{N}$ be the signal of interest, for example, the force readings of the tactile sensors at a given time.
In compressed sensing, the signal is compressed by taking $M$ linear measurements, each of the form,
\[
y_i = \sum_{j=1}^N \measMat_{ij} x_j.
\]
The \emph{measurement vector} $y \in \mR^M$, with components $y_i$, is given by
\begin{equation} \label{cs.eq}
y = \measMat x,
\end{equation}
where $\measMat = [ \measMat_{ij}]$ is the $M \times N$ \emph{measurement matrix}.

If $M < N$, the linear system (\ref{cs.eq}) is under-determined, and thus, in general, it is not possible to recover $x$ from fewer than $N$ measurements.
The theory of compressed sensing shows that, if $x$ is sparse, meaning it has only a few non-zero components, and $\measMat$ satisfies
a property known as the \emph{restricted isometry property}~\cite{cs}, the vector $x$ can be recovered from $M << N$ measurements, on the order of $O(\log N)$ under ideal conditions.
Typically, the signal $x$ is not itself sparse, but may be sparse in some \emph{representation basis}.  Specifically, a signal $x$ is sparse in a basis $\basis \in \mR^{N \times N}$ if
there is a sparse vector $\theta$ such that $x = \basis \theta$.  In this case, the measurement vector can be written as
\[
y = \measMat \basis \theta.
\]
The objective is then to recover $\theta$ from the measurements.
%$x = \basis^{-1} \theta$.
The compressed sensing recovery problem can be formalized as 
\begin{equation} \label{cs1.eq}
 \underset{\theta \in \mR^N}{\text{minimize}}~\| \theta \|_0 ~~~\text{subject to}~\measMat \basis \theta = y
\end{equation}
Here $\| \cdot \|_0$ denotes the $\ell_0$ pseudonorm, \emph{i.e.}, the number of non-zero components.
Once the solution to this problem $\hat{\theta}$ has been obtained, one can recover the original non-sparse signal $x$ as
$x = \basis \hat{\theta}$.
%$x = \basis ^{-1} \hat{\theta}$.

In this work, we consider a variation on the compressed sensing problem where measurements may not be exact.
The measurement vector is given by
\[
y = \measMat \basis \theta + \ev,
\]
where $\ev \in \mR^{M}$ is the measurement noise.  This noise can model both errors in the individual sensor readings as well as in the measurement acquisition.
The sparse vector $\theta$ can be found as the solution of the following optimization problem
\begin{equation} \label {cs2.eq}
 \underset{\theta \in \mR^N}{\text{minimize}}~\textstyle \frac{1}{2}   \displaystyle \| \measMat \basis \theta - y \ \|^2_2 + \lambda \| \theta \|_0,
\end{equation}
where $\lambda$ is a scalar-valued parameter used to tune the level of sparsity of the solution.
It is intractable to solve problem (\ref{cs2.eq}) directly, however, efficient alternate approaches to find $\hat{\theta}$ have been derived~\cite{cs}.

The approach we use in this work is Basis Pursuit Denoising (BPDN). In BPDN, one solves a convex relaxation of (\ref{cs2.eq}), 
\begin{equation} \label {bpdn.eq}
 \underset{\theta \in \mR^N}{\text{minimize}}~\textstyle \frac{1}{2} \displaystyle \| \measMat \basis \theta - y \ \|^2_2 + \lambda \| \theta \|_1.
\end{equation}
Since this problem is convex, it can be solved efficiently using one of many algorithms for convex optimization, as well as variants 
developed specifically for compressed sensing.
It has been shown that BPDN produces optimal or near-optimal sparse solutions in a wide variety of settings~\cite{TW10}.

%\note{Add something about theoretical requirements for $\measMat$, $\basis$, $M$ and $K$.}

\subsection{Compressed Sensing for Tactile Skins} \label{cs4skins.ssec}

To apply compressed sensing to tactile skins, we want to take linear combinations of subsets of taxels' values to generate measurements.  From those measurements, we want to reconstruct the individual taxels' values.  This is done once per time-step and allows us to recover the full data while reducing the number of measurements.

To successfully apply compressed sensing to the tactile sensing problem, we must address the following three components:

\begin{itemize}
\item{\textbf{Identification of a representation basis.}}  We must identify a basis under which the signals (sensor data) from our data sets are sufficiently sparse.  
A greater level of sparsity means that the signals can be accurately reconstructed from fewer measurements.  Further, the representation basis should be universal, 
meaning that it can be used to ``sparsify'' signals in all scenarios.   

\item \textbf{Identification of a sparse measurement matrix that is compatible with the basis and hardware.}  As mentioned above, it is necessary that $\measMat \basis$
satisfy the restricted isometry property. In the tactile sensing application, we have a further restriction that the measurement matrix be sparse. This sparsity is critical as it allows the measurement matrix to be implemented in hardware and reduce the wiring complexity.  This is shown in Figure \ref{measurement.fig}, where in Figure \ref{measurement.fig}a shows each signal with a dedicated connection and Figure \ref{measurement.fig}b shows wiring designed from a compressed sensing measurement matrix where sets of taxels are daisy-chained.  Sparsity is necessary since the non-zero elements in each row indicate the linear combinations (\emph{i.e.}, the taxels with corresponding non-zero values in that row need to be part of the same daisy-chain, such as taxels 1, 2, and 5 in Figure \ref{measurement.fig}b) and the more taxels per linear combination typically corresponds to longer and more complex daisy-chains.  So we seek to minimize this interaction to reduce the amount of wiring. 

\begin{figure}
\includegraphics[width=\linewidth]{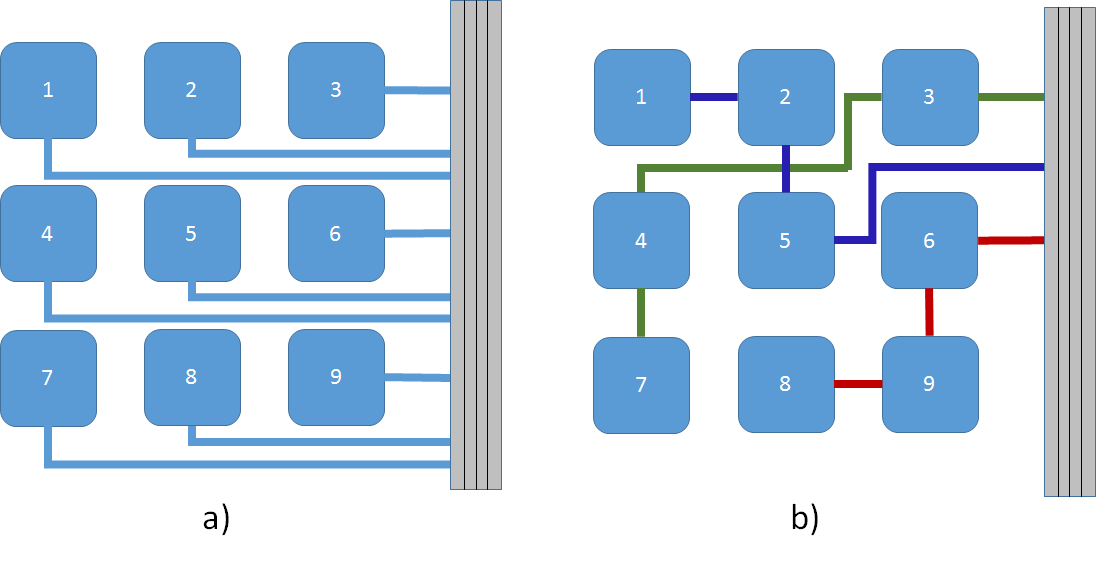}
\caption{Example wiring schematics for a) individual sensor measurements and b) compressed sensing aggregated measurements on a $3 \times 3$ tactile grid.}
\label{measurement.fig}
\end{figure}

\item \textbf{Selection/implementation of a fast signal reconstruction algorithm.} The reconstruction algorithm needs to be able to reconstruct the signal in real time so the signal can be used in control loops.  As reconstruction is a tradeoff between time and accuracy, it also ought to reconstruct the signal at least as a good approximation of the noisy signal, but ideally as a good approximation of the actual signal.
\end{itemize}

In the following section, we describe our approach to each of these three components. 
%Section \ref{sec:rslt} then evaluates the performance of the combined system.

%\note{semi formal definition of tactile problem.}
%\note{desired properties - list (possibly numbered}

%\note{To be implemented in hardware, we want to limit the amount of communication between sensors, (reduce wiring).  This means limited number of non-zero elements in sensing matrix.}

%%%%%%%%%%%%%%%%%%%%%%%%%%%%%%%%%%%%%%%%%%%%%%%%%%%%%%%%%%%%%%%%%%%%%%%%%%%%%%%%
\section{SOLUTION}\label{sec:soln}

\subsection{Representation Basis}
%Daubechies 4 and 2(Haar).  Visualization of sparsity

%\begin{table}[ht]
%\caption{The number of non-zero elements over time for different bases.}
%\begin{center}
%  \begin{tabular}{| c | c | c | c | c | c |}
%    \hline
%    Scenario & $N$ & \multicolumn{2}{c|}{Haar} & \multicolumn{2}{c|}{D4} \\
%    & & mean & max & mean & max\\ \hline
%    LIFT BOX    & 1600 & 108.3 & 266 & 750.1 & 1575 \\ 
%    LIFT WRENCH & 1600 &  92.2 & 388 & 282.2 & 1575 \\
%    DRAG WRENCH & 1600 &  84.9 & 449 & 254.3 & 1575 \\
%    PATH        & 1600 &  30.3 & 266 & 139.8 & 1575 \\
%    LIFT BOX    & 4096 & 189.8 & 263 & 929.1 &  961 \\
%    DRAG WRENCH & 4096 & 240.3 & 670 & 602.7 & 1208 \\
%    \hline
%  \end{tabular}
%\label{tab:BasisSparsity}
%\end{center}
%\end{table}

Historically, tactile processing often used computer vision techniques since the common array pattern of tactile sensor systems corresponds well with image pixel arrays~\cite{dahiya13tactilereview}.
Following this trend, we looked at compressed sensing of images to identify a basis $\basis$ that will render the tactile data signal sufficiently sparse.  
As candidate bases, we considered the Daubechies wavelet transforms, which are similar to the wavelet transforms in JPEG2000.
Daubechies wavelet transforms are a set of transforms where each individual transform is referenced by an even number (\emph{e.g.}, D4) such that larger numbers correspond to more complex transformations~\cite{daubechies88wavelet}.  

We choose to use the D2 wavelet basis as, for all scenarios, the D2 wavelet (also known as the Haar wavelet) basis outperformed the D4 wavelet basis, yielding signals that were indeed sparse.
 This can be seen in Table \ref{tab:BasisSparsity}, which shows the number of non-zero components in the vector $\basis x(t)$, 
both when $\basis$ is the D2 wavelet and when $\basis$ is the D4 wavelet. 
 The table shows the mean and maximum number of non-zero components over all time-steps for each scenario.

 \begin{table}[ht]
\caption{The number of non-zero elements over time for different bases.}
\begin{center}
  \begin{tabular}{| c | c | c | c | c | c |}
    \hline
    Scenario & $N$ & \multicolumn{2}{c|}{D2} & \multicolumn{2}{c|}{D4} \\
    & & mean & max & mean & max\\ \hline
    LIFT BOX    & 1600 & 108.3 & 266 & 750.1 & 1575 \\ 
    LIFT WRENCH & 1600 &  92.2 & 388 & 282.2 & 1575 \\
    DRAG WRENCH & 1600 &  84.9 & 449 & 254.3 & 1575 \\
    PATH        & 1600 &  30.3 & 266 & 139.8 & 1575 \\
    LIFT BOX    & 4096 & 189.8 & 263 & 929.1 &  961 \\
    DRAG WRENCH & 4096 & 240.3 & 670 & 602.7 & 1208 \\
    \hline
  \end{tabular}
\label{tab:BasisSparsity}
\end{center}
\end{table}

One point to note is that, while we represent the signal as a one-dimensional vector, the sensor array is a two-dimensional structure. Thus we use the two-dimensional Haar wavelet for our basis in the same manner that it is used for two-dimensional images.

\subsection{Sensing Matrix}
\begin{figure*}[t]
\includegraphics[width=\linewidth]{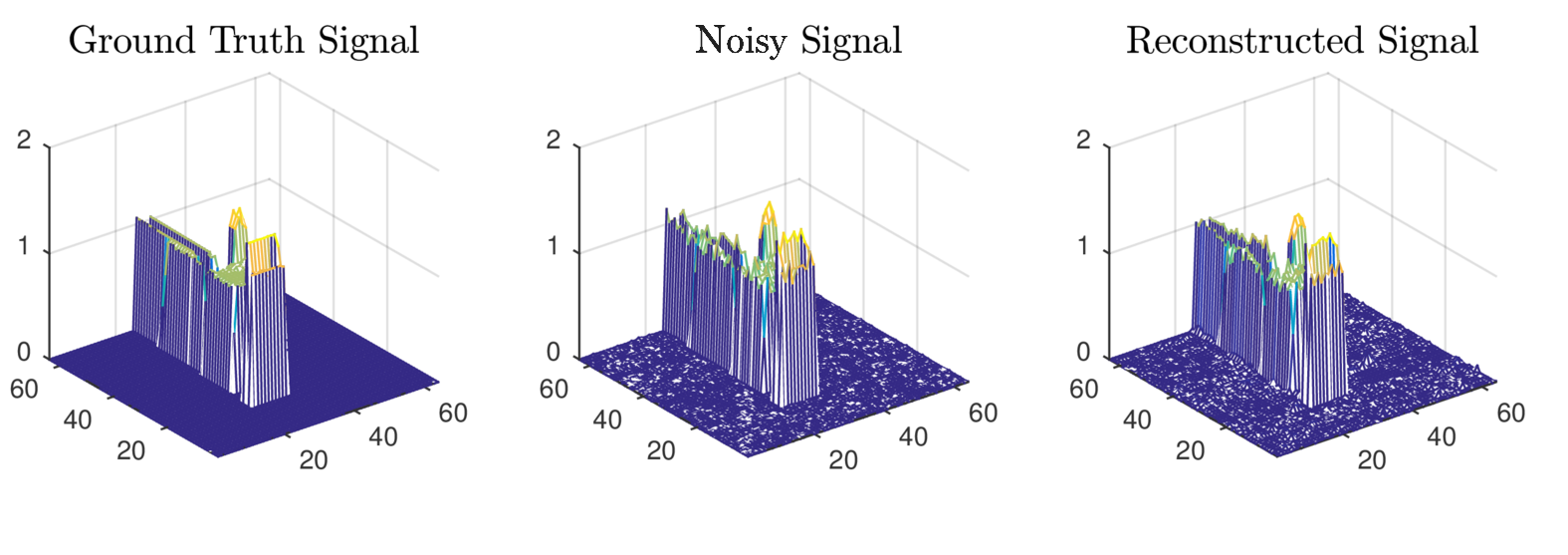}
\caption{Sensor values from DRAG WRENCH ($64 \times 64$) scenario for the ground truth signal (left), noisy signal (middle) and the reconstructed signal (right). The reconstruction was from 1365 measurements using 20 iterations of the FISTA algorithm and has a PSNR of 31.2.}
\label{array.fig}
\end{figure*}

\begin{table*}[ht]
\caption{Quality of Signal Reconstruction with Various Number of Measurements using 30 Iterations}
{\small
\hfill{}
%\caption{Data Results}
%\begin{center}
  \begin{tabular}{| c | c | c | c | c | c |}
    \hline
    Scenario & No. of & \multicolumn{2}{|c|}{Reconstructed Signal}
    & \multicolumn{2}{|c|}{Noisy Signal} \\
    (No. of Elements) & Measurements & Mean PSNR & PSNR Range & Mean PSNR & PSNR Range\\ \hline
    	\multirow{3}{*}{
    	\begin{tabular}{@{}c@{}} LIFT BOX\\(1600) \end{tabular}
    	%\multicolumn{1}{p{2.5cm}}{\centering LIFT BOX\\(1600)}
    	}
    		&  400 & 52.8 & 39.5 - 63.2 & &          \\
		&  533 & 52.5 & 40.9 - 60.8 & 46.5 & 39.1 - 48.7 \\
		&  800 & 52.1 & 41.4 - 57.6 & &                  \\ \hline
  	\multirow{3}{*}{
  		\begin{tabular}{@{}c@{}} LIFT WRENCH\\(1600) \end{tabular}
  	%\multicolumn{1}{p{2.5cm}}{\centering LIFT WRENCH\\(1600)}
  	}
        &  400 & 52.1 & 26.5 - 62.3 & &                  \\
		&  533 & 52.7 & 31.1 - 60.4 & 47.2 & 38.4 - 	48.8 \\
		&  800 & 52.6 & 36.9 - 58.0 & &                  \\ \hline
  	\multirow{3}{*}{
  		\begin{tabular}{@{}c@{}} DRAG WRENCH\\(1600) \end{tabular}
  	%\multicolumn{1}{p{2.5cm}}{\centering DRAG WRENCH\\(1600)}
  	}
        &  400 & 44.7 & 25.3 - 61.9 & &                  \\
		&  533 & 47.7 & 28.5 - 59.8 & 46.4 & 37.4 - 48.7 \\
		&  800 & 48.9 & 34.5 - 58.0 & &                  \\ \hline
  	\multirow{3}{*}{
  		\begin{tabular}{@{}c@{}} PATH\\(1600) \end{tabular}
  	%\multicolumn{1}{p{2.5cm}}{\centering PATH\\(1600)}
  	}
        &  400 & 56.3 & 45.4 - 63.1 & &                  \\
		&  533 & 55.9 & 49.2 - 60.7 & 48.0 & 44.8 - 49.1 \\
		&  800 & 55.0 & 50.3 - 58.4 & &                  \\ \hline
  	\multirow{3}{*}{
  		\begin{tabular}{@{}c@{}} LIFT BOX\\(4096) \end{tabular}
  	%\multicolumn{1}{p{2.5cm}}{\centering LIFT BOX\\(4096)}
  	}
    		& 1024 & 49.8 & 38.6 - 54.9 & &                  \\
		& 1365 & 49.9 & 39.0 - 54.1 & 46.8 & 36.6 - 48.5 \\
		& 2048 & 50.2 & 38.3 - 53.6 & &                  \\ \hline
	\multirow{3}{*}{
		\begin{tabular}{@{}c@{}} DRAG WRENCH\\(4096) \end{tabular}
	%\multicolumn{1}{p{2.5cm}}{\centering DRAG WRENCH\\(4096)}
	}
        	& 1024 & 45.4 & 23.3 - 54.8 & &                  \\
		& 1365 & 46.4 & 29.8 - 53.9 & 46.4 & 36.2 - 48.3 \\
		& 2048 & 48.3 & 37.0 - 53.2 & &                  \\ \hline
        	\hline
  \end{tabular}}
%\label{tab:Results}
%\end{center}
\hfill{}
%\caption{Data Results}
\label{tab:MResults}
\end{table*}

As stated in Section~\ref{sec:cstheory}, it is necessary that the sensing matrix $\measMat$ be chosen so that $\measMat \basis$ satisfies the restricted isometry property.
Unfortunately, it is NP-Hard to verify whether a given matrix $\measMat \basis$ satisfies this property.
It has been shown that if $\measMat$ is a random matrix with entries drawn from a Gaussian distribution, the restricted isometry property is satisfied with high probability~\cite{candes2006compressive}.
While this choice of $\measMat$ is feasible from an algorithmic perspective, such a matrix is dense.  Thus, it is problematic to implement this  measurement approach in hardware due to the issues discussed in Section~\ref{cs4skins.ssec}.

We instead select a sparse measurement matrix that will support data collection and compression with less wiring.
We use the Scrambled Block Hadamard Ensemble (SBHE) developed by Gan et al.~\cite{gan08sbhe}.  It has been shown that for this choice of $\measMat$, the product $\measMat\basis$
behaves like a random Gaussian matrix for a wide variety of bases.
SBHE is a partial block Hadamard transform with randomly permuted columns. It can be represented as 
\begin{equation}
\measMat_H = Q_MWP_N,
\end{equation}
where $W$ is a $N \times N$ block diagonal matrix with each block a $B \times B$ Hadamard matrix, $P_N$ is the permutation matrix, which randomly reorders the $N$ columns of $W$, and $Q_M$ uniformly at random selects $M$ rows of $WP_N$.  As the non-zero elements come from the $B \times B$ Hadamard blocks in $W$, $B$ determines the number of sensors connected together for each measurement element. From a wiring standpoint, we want to minimize $B$ and, similar to Gan et al., we found $B = 32$ worked well for our evaluations.  

SBHE also has two additional benefits.  First, all non-zero elements are either $+1$ or $-1$.  This simplifies the hardware as taxels only need the additional hardware to negate their values instead of various hardware for different gains for each measurement, as would be the case for the random Gaussian matrix.
Second, the taxels are divided into disjoint measurement subsets that each give multiple measurements.  This means for any two measurements, either the whole subset of taxels or none of the taxels are used in both measurements. 
This is a benefit as no wires are needed between the subsets, allowing for very distinct daisy-chains. This property also has the potential for the measurements to share wires.

\subsection{Reconstruction Algorithm}

To solve problem (\ref{bpdn.eq}), we use an iterative, gradient-like method called Fast Iterative Shrinkage-Thresholding Algorithm (FISTA)~\cite{BT09}.
We selected this algorithm because of its fast convergence; FISTA converges at a rate of $O(1/k^2)$, where $k$ is the iteration number, whereas standard gradient methods converge 
at a rate of $O(1/k)$. Every time-step we use FISTA to reconstruct the signal from the measurements taken in that time-step.  As the signal typically does not change much between time-steps, we use the previous time-step's reconstructed signal as the initial estimate for the current signal and use FISTA to refine the estimate into a solution.
This warm-start led to faster convergence to the solution.
FISTA is also amenable to hardware-based implementations, which can further accelerate its performance.
We exploit this feature in our implementation.
%I do not know what these last two sentences mean, unless I am referring to GPU

%%%%%%%%%%%%%%%%%%%%%%%%%%%%%%%%%%%%%%%%%%%%%%%%%%%%%%%%%%%%%%%%%%%%%%%%%%%%%%%%

\section{SIMULATION RESULTS}\label{sec:rslt}

\subsection{Algorithm Implementation}
We implemented FISTA in Matlab 2015a, utilizing Matlab's GPU capabilities to speed up the algorithm's execution.   
The resulting code required approximately 1ms per iteration (see Table \ref{tab:IterResults}). 
We ran the algorithm for a constant number of iterations for each compressed sensing problem as it standardized the reconstruction time, and
 checking for other possible exit criteria greatly reduced performance. 
We performed simulations for several values of $\lambda$ (see (\ref{bpdn.eq})) and determined that
$\lambda = 0.1$ worked well for all scenarios.  We used this parameter value in all simulations.

We noted that the reconstructed signal returned from FISTA sometimes contained negative values, but as we know this is not a feasible signal, we set those elements to 0.

Our experiments were run on a machine using Linux Ubuntu 14.04.  The machine contained an Intel Core i7 CPU (8 cores at 3.07~GHz) and a single GeForce GTX 480 GPU.

\subsection{Noise Model}
We added noise to the tactile data before performing the measurements.  
Our noise model was a normal distribution with zero mean.  The model had a standard deviation of 0.0628 Newtons (5 percent) at the sensor range midpoint and tapered towards the ends of the sensor range.  All values beyond the sensor range were modified to the corresponding boundary value.

\subsection{Results}

To test our method, we applied our measurement matrix $\measMat_H$ to our scenarios and then reconstructed the signal with FISTA.  An illustrative example of the sensor data and reconstruction is shown in
Figure \ref{array.fig}.  The figure shows the three signals for a single time-step: the ground truth signal (\emph{i.e.}, the SkinSim filtered signal) of the object contact, the noisy signal (\emph{i.e.}, the ground truth signal plus our noise model) that we compress for our measurements, and the reconstructed signal (\emph{i.e.}, the restored signal from FISTA). 
From the figure it is evident that the reconstructed signal closely matches the shape and magnitude of the original signal, demonstrating our method  gives reasonable results.
% I wouldn't say this because as a reader, my question is, why did you choose to show a poor example?
% especially considering this time step evaluated poorly compared to the majority of our data (see Tables~\ref{tab:IterResults} and \ref{tab:MResults}).

To evaluate the quality of our reconstruction, for each time-step we measured the peak signal-to-noise ratio (PSNR) of the reconstructed signal and compared it to the PSNR of the noisy signal (larger PSNRs are better).

We first evaluated the impact of the number of measurements on the reconstruction quality. We varied the number of measurements to $\frac{N}{4},\frac{N}{3}$ and $\frac{N}{2}$, where $N$ is the number of taxels.  For these experiments, the number of FISTA iterations was set to 30.  We selected 30 iterations after tracking the number of iterations FISTA ran when using various stopping criteria based on convergence (\emph{e.g.}, the change in the solution between iterations was below a threshold, solution was within a threshold of the ground truth, etc.) and found 30 iterations was typically more than enough.  
The mean and range of the PSNR over time are shown in Table \ref{tab:MResults} for each scenario, along with the PSNR of the noisy signal.  
The table shows that for the vast majority of the reconstructions, the mean PSNR exceeded that of the noisy signal.  
This means that the reconstructed signal more accurately estimated the ground truth than the noisy signal.
In addition, for most scenarios, the reconstructed mean PSNR was similar for all three measurement sizes, though the PSNR range was typically larger for fewer measurements.  This means that while more measurements made the reconstruction quality more consistent, on average the PSNR were independent of the number of measurements.
The scenarios that have mean PSNR less than the noisy signal both used only $\frac{N}{4}$ measurements.
This indicates that $\frac{N}{4}$ measurements can be comparable with the noisy signal accuracy only for some scenarios.

\begin{figure}
\begin{subfigure}{.45\textwidth}
	\includegraphics[width = \textwidth]{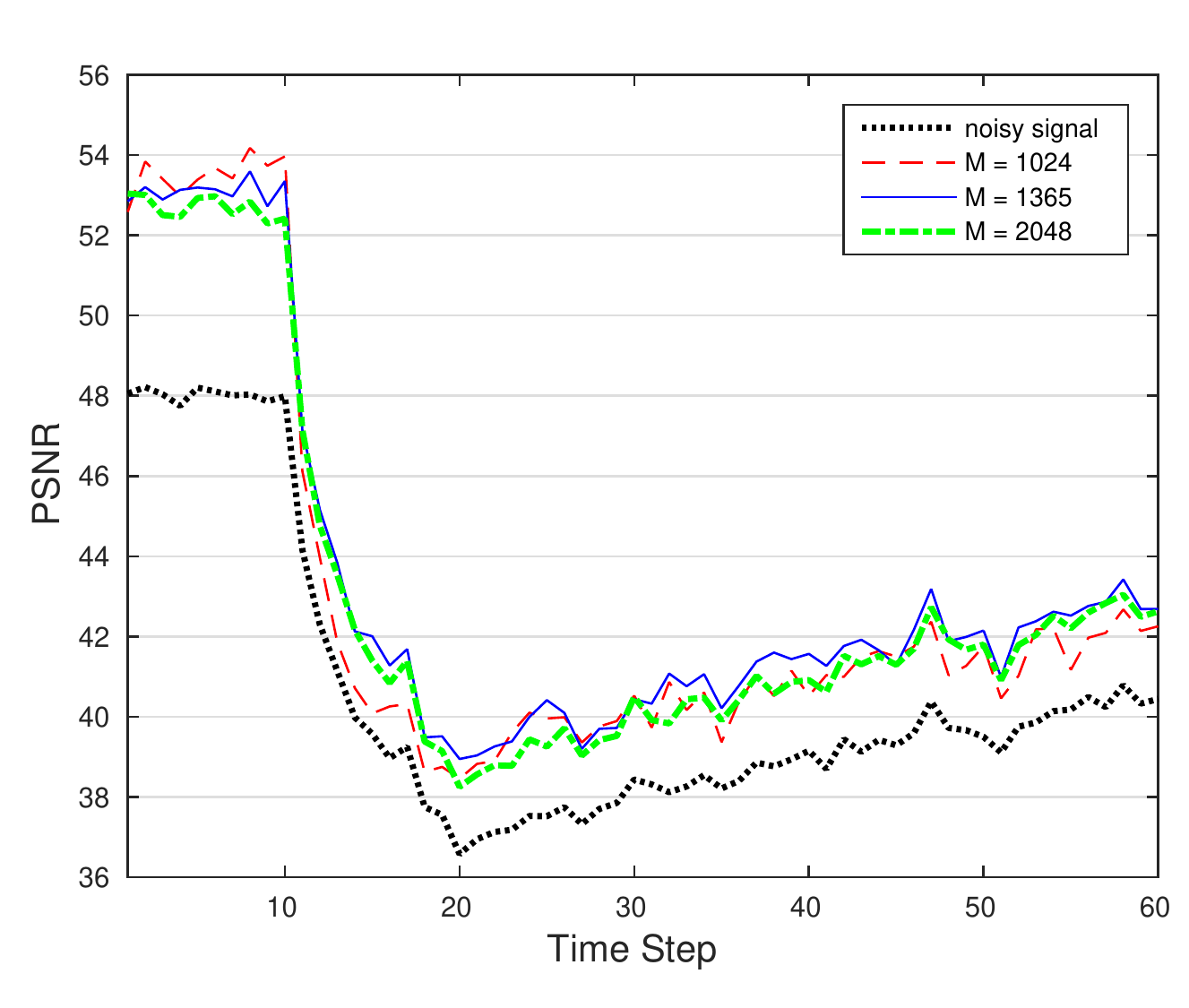}
	\caption{Representative Case Time Sequence - LIFT BOX $64\times64$}
	\label{bm.fig}
\end{subfigure}
\begin{subfigure}{.45\textwidth}
	\includegraphics[width = \textwidth]{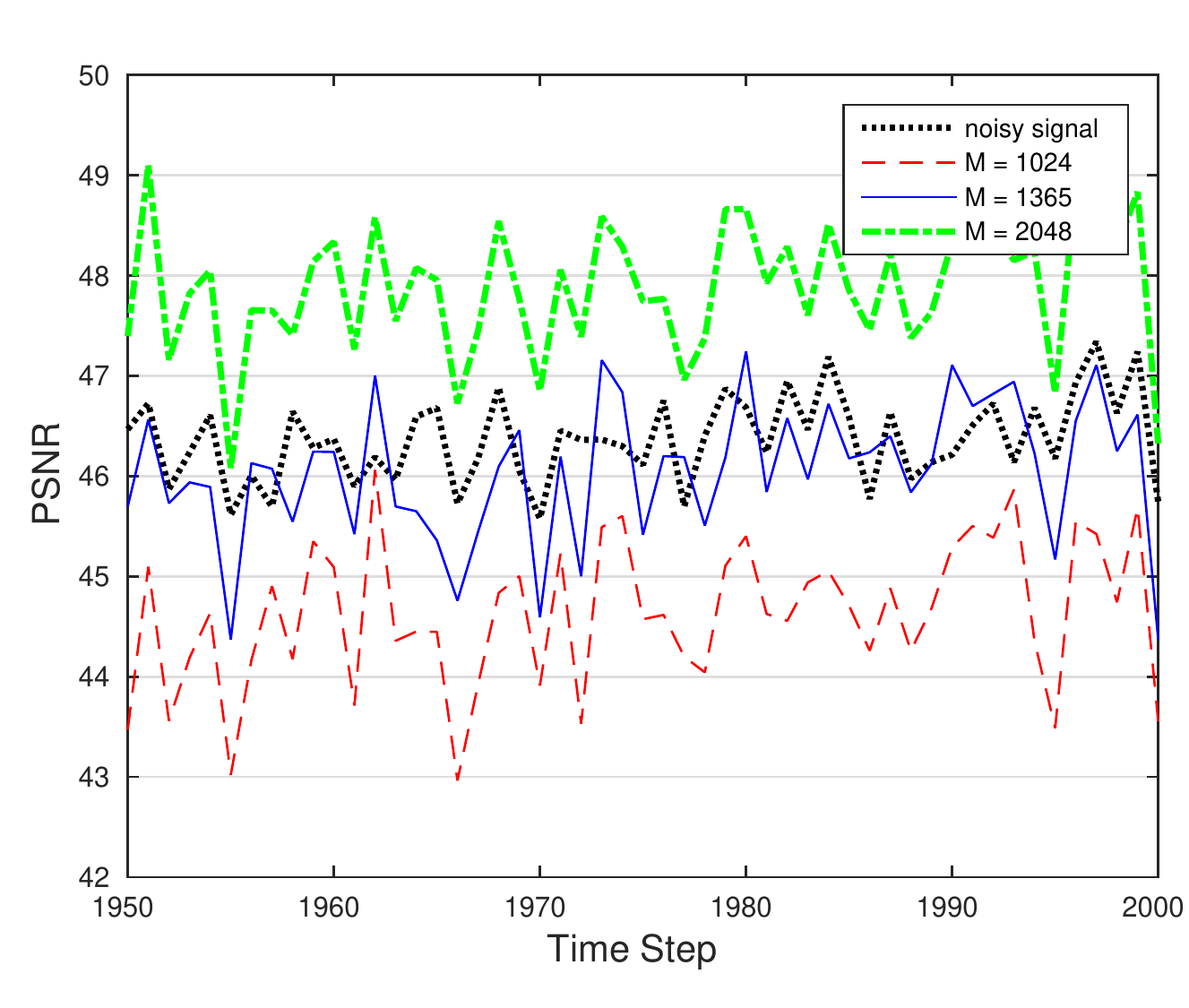}
	\caption{Worst Case Time Sequence - DRAG WRENCH $64\times64$}
	\label{wm.fig}
\end{subfigure}
\caption{Representative time sequences of PSNR of the reconstructed signal for the different numbers of measurements ($\frac{N}{4},\frac{N}{3}, \frac{N}{2}$) as well as the PSNR of the noisy signal. (a) is representative of most scenarios as the reconstructed signal outperforms the noisy signal while (b) is the worst-case scenario where the reconstruction signal only out performs the noisy signal for large numbers of measurements.}
\label{m.fig}
\end{figure}

The lower end of the PSNR range values occurred during the first few time-steps of contact between the object and sensor pad.  This visually results in a dip in the PSNR plot as seen in Figure \ref{bm.fig}.  The reason for this dip is our noise model produces more noise at the middle of the sensor range than at the ends, and prior to contact the sensor values are at the low end of the range and during first contact, the sensor values are closer to the middle of the sensor range.

Figure \ref{m.fig} shows representative time sequences of how the number of measurements affects the reconstruction using the LIFT BOX and DRAG WRENCH scenarios, both using the $64 \times 64$ array.
The results for the LIFT BOX scenario in Figure \ref{bm.fig} are fairly representative of most of the scenarios. The reconstructed PSNR does not vary much for different numbers of measurements and is larger than the noisy PSNR.
Figure \ref{wm.fig} shows a representative snapshot of the worst performing scenarios, the DRAG WRENCH scenarios.
The PSNR for $\frac{N}{4}$ measurements was consistently below that of the noisy signal,
the PSNR for $\frac{N}{3}$ measurements was similar to that of the noisy signal, and for $\frac{N}{2}$ measurements,
the PSNR was better than that of the noisy signal.  
This figure shows that, in the worst case, using $\frac{N}{3}$ measurements gives accuracy on par with the noisy signal.
These results show that our measurement approach compresses the signal up to a third of its original size,
with the ability to recover the original signal typically with higher quality than the noisy signal.

We next looked at the impact of the number of iterations on the reconstruction accuracy, which is an indication of the reconstruction speed.
We ran FISTA for 10, 20, and 30 iterations and evaluated the quality of the reconstructed signals.
We used $\frac{N}{3}$ measurements in all cases.
The results are presented in Table \ref{tab:IterResults}.
The average PSNR for each scenario was unaffected by the number of iterations, but the minimum PSNR varied for most scenarios, especially for 10 iterations.  
This is especially of interest as the few time-steps that produced the lower end of the PSNRs' range were the only time steps it was challenging for the reconstructed signal to be comparable to the noisy signal.

%\begin{center}
\begin{table*}[ht]
\caption{Quality and Time of Signal Reconstruction with Various Number of Iterations using $\frac{N}{3}$ Measurements}
{\small
\hfill{}
%\caption{Data Results}
%\begin{center}
  \begin{tabular}{| c | c | c | c | c | c | c | c |}
    \hline
    Scenario & No. of & No. of & Mean & \multicolumn{2}{|c|}{Reconstructed Signal}
    & \multicolumn{2}{|c|}{Noisy Signal} \\
    (No. of Elements) & Measurements & Iterations & Time & Mean PSNR & PSNR Range & Mean PSNR & PSNR Range\\ \hline
    	\multirow{3}{*}{
    	\begin{tabular}{@{}c@{}} LIFT BOX\\(1600) \end{tabular}
    	%\multicolumn{1}{p{2.5cm}}{\centering LIFT BOX\\(1600)}
    	}
    		&     & 10 & 10.8 & 52.5 & 37.0 - 60.8 &  &  \\
    		& 533 & 20 & 20.7 & 52.5 & 40.9 - 60.8 & 46.5 & 39.1 - 48.7 \\   		
        	&     & 30 & 30.9 & 52.5 & 40.9 - 60.8 &  &  \\ \hline
  	\multirow{3}{*}{
  		\begin{tabular}{@{}c@{}} LIFT WRENCH\\(1600) \end{tabular}
  	%\multicolumn{1}{p{2.5cm}}{\centering LIFT WRENCH\\(1600)}
  	}
        	&     & 10 & 10.6 & 52.7 & 26.7 - 60.4 &  &  \\
        	& 533 & 20 & 20.6 & 52.7 & 30.7 - 60.4 & 47.2 & 38.4 - 48.8 \\
        	&     & 30 & 30.1 & 52.7 & 31.1 - 60.4 &  &  \\ \hline
  	\multirow{3}{*}{
  		\begin{tabular}{@{}c@{}} DRAG WRENCH\\(1600) \end{tabular}
  	%\multicolumn{1}{p{2.5cm}}{\centering DRAG WRENCH\\(1600)}
  	}
        	&     & 10 & 10.7 & 47.7 & 26.3 - 59.8 & &  \\
        	& 533 & 20 & 20.7 & 47.7 & 28.4 - 59.8 & 46.4 & 37.4 - 48.7 \\
        	&     & 30 & 28.1 & 47.7 & 28.5 - 59.8 &  &  \\ \hline
  	\multirow{3}{*}{
  		\begin{tabular}{@{}c@{}} PATH\\(1600) \end{tabular}
  	%\multicolumn{1}{p{2.5cm}}{\centering PATH\\(1600)}
  	}
        	&     & 10 & 10.3 & 55.9 & 49.2 - 60.7 &  &   \\
        	& 533 & 20 & 18.1 & 55.9 & 49.1 - 60.7 & 48.0 &  44.8 - 49.1 \\
        	&     & 30 & 28.8 & 55.9 & 49.2 - 60.7 &  &   \\ \hline
  	\multirow{3}{*}{
  		\begin{tabular}{@{}c@{}} LIFT BOX\\(4096) \end{tabular}
  	%\multicolumn{1}{p{2.5cm}}{\centering LIFT BOX\\(4096)}
  	}
    		&      & 10 &  8.1 & 49.9 & 32.1 - 54.1 &  &  \\
    		& 1365 & 20 & 21.6 & 49.9 & 38.9 - 54.1 & 46.8 & 36.6 - 48.5 \\    		
        	&      & 30 & 25.1 & 49.9 & 39.0 - 54.1 &  &  \\ \hline
	\multirow{3}{*}{
		\begin{tabular}{@{}c@{}} DRAG WRENCH\\(4096) \end{tabular}
	%\multicolumn{1}{p{2.5cm}}{\centering DRAG WRENCH\\(4096)}
	}
        	&      & 10 &  8.0 & 46.4 & 23.1 - 53.9 &  &  \\
        	& 1365 & 20 & 17.0 & 46.4 & 29.3 - 53.9 & 46.4 & 36.2 - 48.3 \\
        	&      & 30 & 31.7 & 46.4 & 29.8 - 53.9 &  &  \\ \hline
        	\hline
  \end{tabular}}
%\label{tab:Results}
%\end{center}
\hfill{}
%\caption{Data Results}
\label{tab:IterResults}
\end{table*}
%\end{center}
\begin{figure*}
\begin{subfigure}{.5\textwidth}
	\includegraphics[width = .95\textwidth]{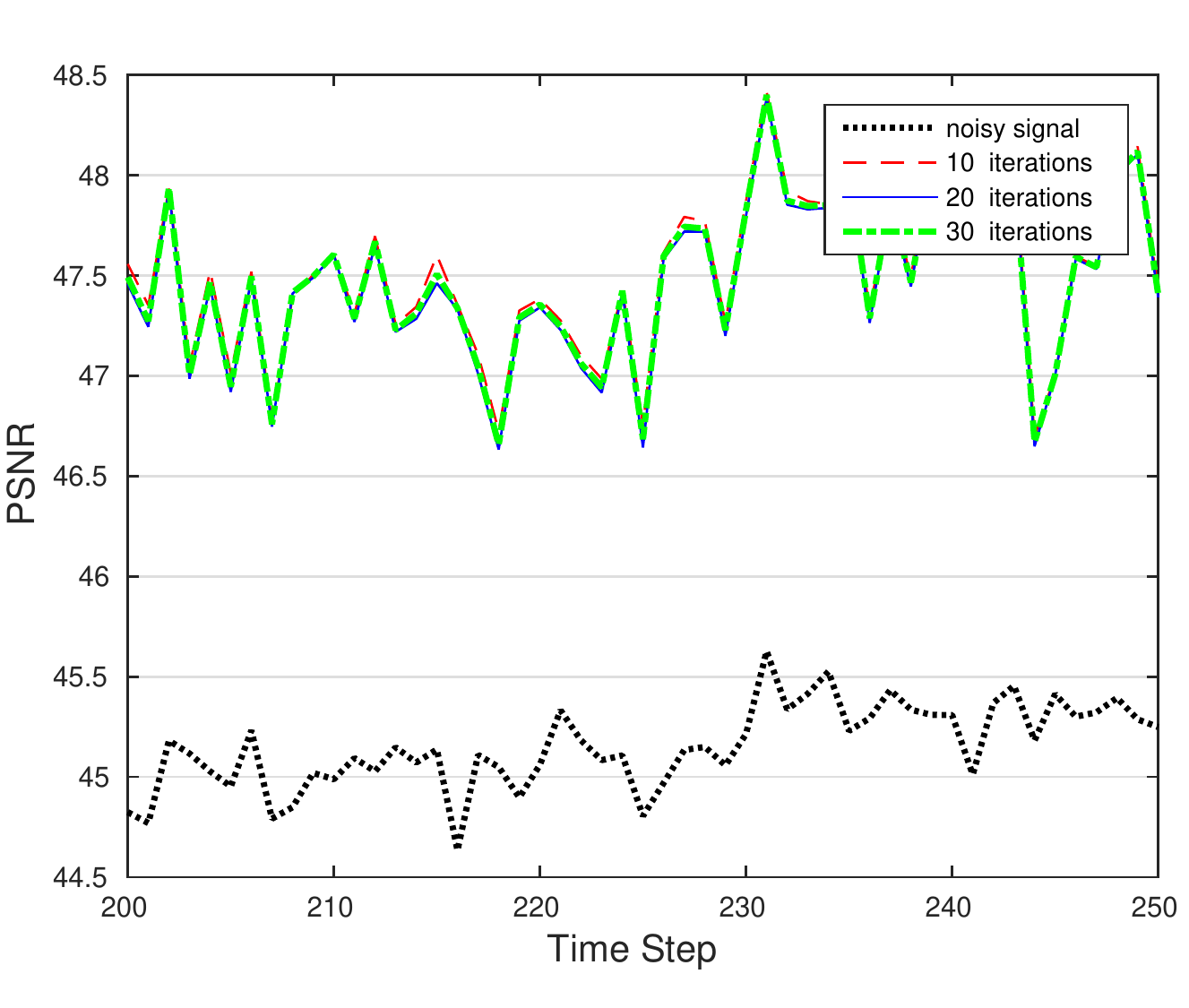}
	\caption{Representative Case Time Sequence - LIFT BOX $64\times64$}
	\label{biter.fig}
\end{subfigure}
\begin{subfigure}{.5\textwidth}
	\includegraphics[width = .95\textwidth]{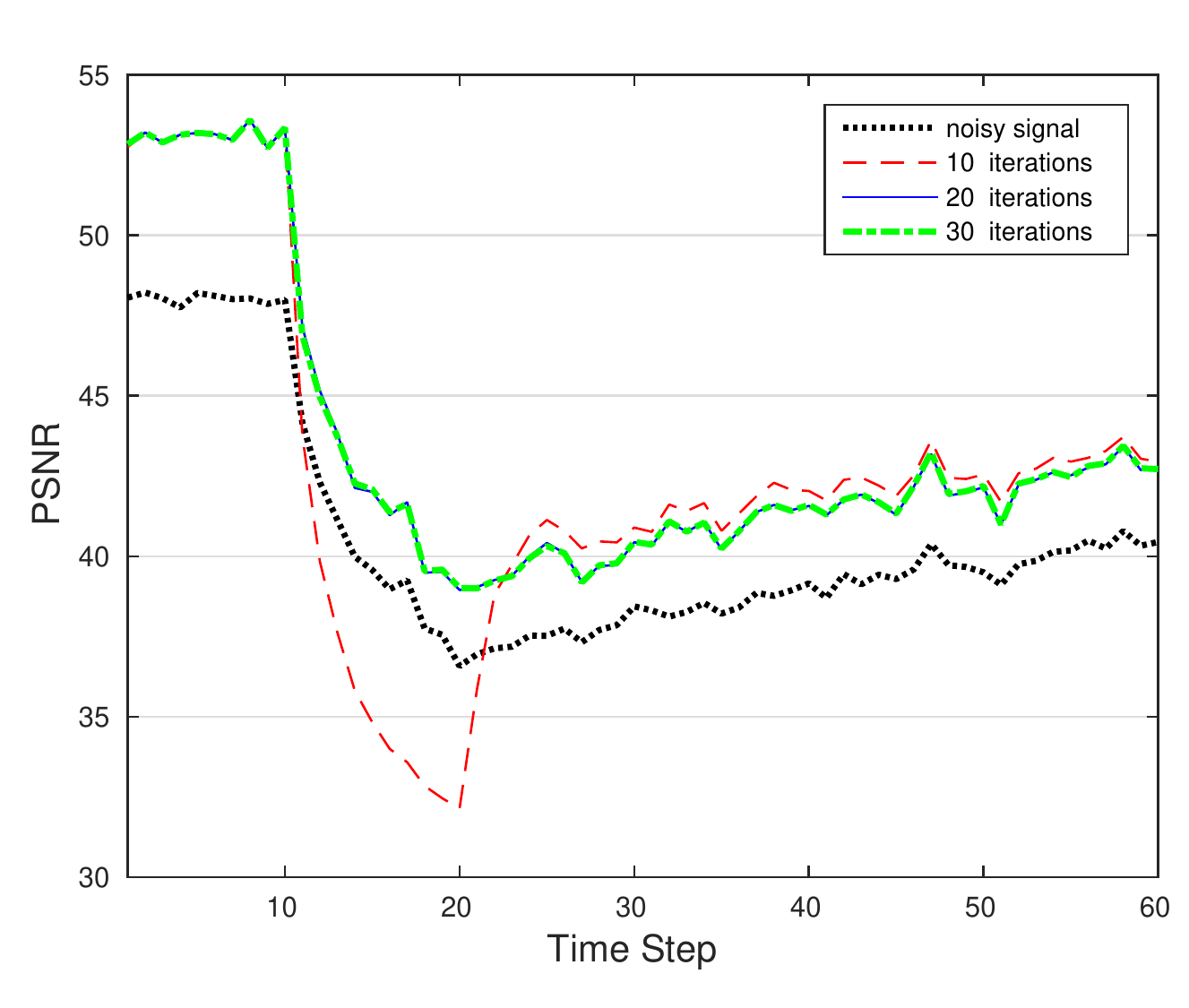}
	\caption{Worst Case Time Sequence - LIFT BOX $64\times64$}
	\label{witer.fig}
\end{subfigure}
\caption{Both plots are representative time sequences of PSNR for LIFT BOX $64\times64$ scenario with 10, 20, and 30 iterations and the PSNR of the noisy signal. (a) is a representative time sequence for the majority of time-steps across the scenarios, while (b) is the worst-case time-steps. }
\label{iter.fig}
\end{figure*}
To further illustrate how the number of iterations affects the PSNR, Figure \ref{iter.fig} shows representative time sequences for the parts of the reconstructed signal using scenario LIFT BOX $64 \times 64$.  Both Figures \ref{biter.fig} and \ref{witer.fig} show the PSNR for reconstructed signal after 10, 20, and 30 FISTA iterations, along with the PSNR for the noisy signal for comparison. Figure \ref{biter.fig} is a snapshot of the majority of the time sequences where the ground truth signal changes slowly, while Figure \ref{witer.fig} covers the time sequence of the first contact. Where the ground truth signal changes slowly, which is the majority of the time for all scenarios, the reconstructed signals' PSNR plots are nearly indistinguishable for the different iterations, which is shown in Figure \ref{biter.fig}. 
Figure \ref{witer.fig} similarly shows the PSNR plots are nearly identical, except that for a few time-steps the PSNR for 10 iterations was significantly lower than for 20 and 30 iterations.  These time-steps occurred when the object initially made contact with the sensors, or in other words, the deviation between the PSNR of 10 iterations and the other values occurred at sudden, significant signal changes.  The drastic signal change requires more iterations to converge as the warm-start is no longer a good initial estimate, and Figure \ref{witer.fig}, along with Table \ref{tab:IterResults}, shows 10 iterations is not enough.  The PSNR stabilizes again due to the aggregation of FISTA iterations, which occurs due to the warm-starts. While for some applications, the loss in quality that 10 iterations experienced might be acceptable, this is not always the case.  Significant signal changes often indicate significant changes in the robot state and thus, an accurate reconstruction may be important for generating the correct response. In this case, using 20 iterations is preferable.

%\note{We don't actually have to pick 20 iterations for anything.  We are just illustrating the tradeoffs between 10, 20, and 30.} 
%Therefore we use 20 iterations for our system speed as it is the rate we can consistently reconstruct the signal up to and exceeding the noisy signal, which we know (see Table \ref{tab:IterResults}) takes approximately 20ms.

%\addtolength{\textheight}{-5cm}   % This command serves to balance the column lengths

%%%%%%%%%%%%%%%%%%%%%%%%%%%%%%%%%%%%%%%%%%%%%%%%%%%%%%%%%%%%%%%%%%%%%%%%%%%%%%%%
\section{CONCLUSION AND FUTURE WORK}\label{sec:concl}

\subsection{Conclusion}
We have proposed an approach to tactile data acquisition using compressed sensing.
In our approach, tactile data is compressed in hardware before transmission to the central processing unit.
Then, the fully sensed data signal is reconstructed from the compressed data.

In this work, we have used standard (\emph{i.e.}, pre-existing and non-optimized) tools of compressed sensing, and we still achieved full signal reconstruction of 4096 taxels at a rate of 50~hz, 
which is on the same order of magnitude as the sensors presented by Schmitz et al.~\cite{schmitz11icub}, the sensors used for the system with the most taxels.
Further, our system was able to compress the signal to a third of its original size and produce a higher quality signal than the noisy signal a vast majority of the time.  
Additionally, we have discussed how compressed sensing may provide guidance for new wiring techniques with potential to reduce the wiring complexity.  Using daisy-chain techniques, wiring could be reduced on the same order as the data compression. We were able to reduce the number of measurements by a factor of three, but the theoretical limit for compressed sensing is to have  $O(\log N)$ measurements. 
From this initial investigation,  we conclude that compressed sensing is a feasible approach for tactile data acquisition and worth further exploration.
We believe this paradigm can open the door to larger-scale tactile systems as well as faster data acquisition.

\addtolength{\textheight}{-.1cm}   % This command serves to balance the column lengths

\subsection{Future Work}
We plan to continue to develop our system so as to meet the suggested requirements mentioned in the introduction by exploring additional bases and measurement matrices.  
Not only will we look for representation bases that give even more sparsity in the tactile data, which is a key component for achieving $O(\log N)$ measurements, but we also will explore bases that work with different tactile sensor arrangements (\emph{e.g.}, non-planar). 
Part of our future work in exploring measurement matrices will be to consider the constraint that sensors that interact during measurement be in close proximity. 
This will further simplify the wiring arrangement.
Finally, we will investigate alternatives and enhancements to our reconstruction algorithm to improve reconstruction speed.

%Other future work is to expand into processing the data.  Particularly of interest is distributed processing methods and researching the feasibility of utilizing the compressed data, as this would great improve sampling time.

%%%%%%%%%%%%%%%%%%%%%%%%%%%%%%%%%%%%%%%%%%%%%%%%%%%%%%%%%%%%%%%%%%%%%%%%%%%%%%%%
%\section{ACKNOWLEDGMENTS}
% The acknowledgements should be in the \thanks after the authors' names
%
%The authors gratefully acknowledge the contribution of National Research Organization and reviewers' comments.

%%%%%%%%%%%%%%%%%%%%%%%%%%%%%%%%%%%%%%%%%%%%%%%%%%%%%%%%%%%%%%%%%%%%%%%%%%%%%%%%

\bibliographystyle{IEEEtran}
\bibliography{main}

\begin{thebibliography}{10}
\providecommand{\url}[1]{#1}
\csname url@rmstyle\endcsname
\providecommand{\newblock}{\relax}
\providecommand{\bibinfo}[2]{#2}
\providecommand\BIBentrySTDinterwordspacing{\spaceskip=0pt\relax}
\providecommand\BIBentryALTinterwordstretchfactor{4}
\providecommand\BIBentryALTinterwordspacing{\spaceskip=\fontdimen2\font plus
\BIBentryALTinterwordstretchfactor\fontdimen3\font minus
  \fontdimen4\font\relax}
\providecommand\BIBforeignlanguage[2]{{%
\expandafter\ifx\csname l@#1\endcsname\relax
\typeout{** WARNING: IEEEtran.bst: No hyphenation pattern has been}%
\typeout{** loaded for the language `#1'. Using the pattern for}%
\typeout{** the default language instead.}%
\else
\language=\csname l@#1\endcsname
\fi
#2}}

\bibitem{nri}
``National robotics initiative, program solicitation,''
  \url{http://www.nsf.gov/pubs/2015/nsf15505/nsf15505.htm}, [accessed
  2015-09-01].

\bibitem{box}
Y.~Ohmura and Y.~Kuniyoshi, ``Humanoid robot which can lift a 30kg box by whole
  body contact and tactile feedback,'' in \emph{Proc. IEEE/RSJ Int. Conf.
  Intell. Robots Syst.}, 2007, pp. 1136--1141.

\bibitem{dahiya10tactile}
R.~Dahiya, G.~Metta, M.~Valle, and G.~Sandini, ``Tactile sensing: From humans
  to humanoids,'' \emph{Robotics, IEEE Transactions on}, vol.~26, no.~1, pp.
  1--20, Feb 2010.

\bibitem{icub}
A.~Parmiggiani, M.~Maggiali, L.~Natale, F.~Nori, A.~Schmitz, N.~Tsagarakis,
  J.~S. Victor, F.~Becchi, G.~Sandini, and G.~Metta, ``The design of the icub
  humanoid robot,'' \emph{Int. Jour. of Humanoid Robots}, 2011.

\bibitem{ohmura06sensor}
Y.~Ohmura, Y.~Kuniyoshi, and A.~Nagakubo, ``Conformable and scalable tactile
  sensor skin for curved surfaces,'' in \emph{Proc. IEEE Int. Conf. Robotics
  and Automation (ICRA)}, May 2006, pp. 1348--1353.

\bibitem{schmitz11icub}
A.~Schmitz, P.~Maiolino, M.~Maggiali, L.~Natale, G.~Cannata, and G.~Metta,
  ``Methods and technologies for the implementation of large-scale robot
  tactile sensors,'' \emph{Robotics, IEEE Transactions on}, vol.~27, no.~3, pp.
  389--400, June 2011.

\bibitem{mitt11hex}
P.~Mittendorfer and G.~Cheng, ``Humanoid multimodal tactile-sensing modules,''
  \emph{Robotics, IEEE Transactions on}, vol.~27, no.~3, pp. 401--410, June
  2011.

\bibitem{riman}
T.~Mukai, M.~Onishi, T.~Odashina, S.~Hirano, and Z.~Luo, ``Development of the
  tactile sensor system of a human-interactive robot ``{RI-MAN}'','' \emph{IEEE
  Trans. Robot.}, vol.~24, no.~2, pp. 505--512, Apr 2008.

\bibitem{cs}
M.~Davenport, M.~Duarte, Y.~Eldar, and G.~Kutynoik, \emph{Compressed Sensing:
  Theory and Applications}.\hskip 1em plus 0.5em minus 0.4em\relax Cambridge
  University Press, 2012.

\bibitem{gazebo}
``Gazebo,'' \url{http://gazebosim.org}, [accessed 2015-09-13].

\bibitem{habib2014skinsim}
A.~Habib, I.~Ranatunga, K.~Shook, and D.~O. Popa, ``Skinsim: A simulation
  environment for multimodal robot skin,'' in \emph{Proc. {IEEE} Conf.
  Automation Science and Engineering (CASE)}, 2014, pp. 1226--1231.

\bibitem{TW10}
J.~Tropp and S.~J. Wright, ``Computational methods for sparse solution of
  linear inverse problems,'' \emph{Proc. {IEEE}}, vol.~98, no.~6, pp. 948--958,
  2010.

\bibitem{dahiya13tactilereview}
R.~Dahiya, P.~Mittendorfer, M.~Valle, G.~Cheng, and V.~Lumelsky, ``Directions
  toward effective utilization of tactile skin: A review,'' \emph{Sensors
  Journal, IEEE}, vol.~13, no.~11, pp. 4121--4138, Nov 2013.

\bibitem{daubechies88wavelet}
I.~Daubechies, ``Orthonormal bases of compactly supported wavelets,''
  \emph{Communications on Pure and Applied Mathematics}, vol.~41, no.~7, pp.
  909--996, 1988.

\bibitem{candes2006compressive}
E.~J. Cand{\`e}s \emph{et~al.}, ``Compressive sampling,'' in \emph{Proc.
  International congress of mathematicians}, vol.~3, 2006, pp. 1433--1452.

\bibitem{gan08sbhe}
L.~Gan, T.~Do, and T.~Tran, ``Fast compressive imaging using scrambled block
  hadamard ensemble,'' in \emph{Signal Processing Conference, 2008 16th
  European}, Aug 2008, pp. 1--5.

\bibitem{BT09}
A.~Beck and M.~Teboulle, ``A fast iterative shrinkage-thresholding algorithm
  for linear inverse problems,'' \emph{SIAM journal on imaging sciences},
  vol.~2, no.~1, pp. 183--202, 2009.

\end{thebibliography}

\end{document}